\documentclass{article} 
\usepackage{nips15submit_e,times}
\usepackage{hyperref}
\usepackage{url}

\usepackage{times}
\usepackage{graphicx} 
\usepackage{subfigure} 

\usepackage{algorithm}
\usepackage{algorithmic}

\usepackage{amssymb, amsmath}
\usepackage{amsthm}
\usepackage{booktabs}  
\usepackage{color}
\usepackage{mathtools}
\usepackage{multirow}
\usepackage{tabularx}

\usepackage{epstopdf}

\newcommand{\dw}{\nabla_{\w}}
\newcommand{\fhat}{\hat{f}}
\newcommand{\Jhat}{\hat{J}}

\newcommand{\mutil}{\tilde{\nu}}

\newcommand{\sqnorm}[1]{\left|\left|#1\right|\right|^2} 

\newcommand{\vt}{{\bf v}_t}

\newcommand{\w}{{\bf w}}

\newcommand{\wstar}{\w^*}
\newcommand{\wtil}{{\tilde{\bf w}}}
\newcommand{\x}{{\bf x}}

\newcommand{\yhat}{\hat{y}}

\newcommand{\A}{{\mathcal A}}
\newcommand{\B}{{\mathcal B}}
\newcommand{\C}{{\mathcal C}}
\newcommand{\D}{{\mathcal D}}
\newcommand{\E}{{\mathbb E}}
\newcommand{\F}{{\mathcal F}}

\newcommand{\INDSTATE}[1]{\STATE \hspace{2mm} #1}

\newtheorem{mylemma}{Lemma}
\newtheorem{mytheorem}{Theorem}
\newtheorem{myrem}{Remark}

\newcolumntype{S}{>{\centering\arraybackslash} m{.10\linewidth} }
\newcolumntype{T}{>{\centering\arraybackslash} m{.30\linewidth} }
\newcolumntype{U}{>{\centering\arraybackslash} m{.08\linewidth} }
\newcolumntype{V}{>{\centering\arraybackslash} m{.37\linewidth} }
\newcolumntype{W}{>{\centering\arraybackslash} m{.1\linewidth} }

\newcommand{\comment}[1]{}

\title{A Variance Reduced Stochastic Newton Method}

\author{
Aurelien Lucchi \quad\quad Brian McWilliams \quad\quad Thomas Hofmann\\
Department of Computer Science, ETH Z\"urich \\
\scriptsize{\texttt{\{aurelien.lucchi, brian.mcwilliams, thomas.hofmann \} @inf.ethz.ch}} \\
}

\nipsfinalcopy 

\begin{document}

\maketitle

\begin{abstract} 

Quasi-Newton methods are widely used in practise for convex loss minimization problems. These methods exhibit good empirical performance on a wide variety of tasks and enjoy super-linear convergence to the optimal solution. For large-scale learning problems, stochastic Quasi-Newton methods have been recently proposed. However, these typically only achieve sub-linear convergence rates and have not been shown to consistently perform well in practice since noisy Hessian approximations can exacerbate the effect of high-variance stochastic gradient estimates.   
In this work we propose {\sc Vite}, a novel stochastic Quasi-Newton algorithm that uses an existing first-order
technique to reduce this variance. Without exploiting the specific form of the approximate Hessian, we show that {\sc Vite} reaches the optimum at a geometric rate with a constant step-size when dealing
with smooth strongly convex functions. Empirically, we demonstrate improvements over existing stochastic Quasi-Newton and variance reduced stochastic gradient methods.

\end{abstract}

\section{Introduction}

We consider the problem of optimizing a function expressed as an expectation
over a set of data-dependent functions. Stochastic gradient
descent (SGD) has become the method of choice for such tasks as it only requires computing stochastic gradients over
a small subset of datapoints \cite{bottou2010,
  shalev2011}. The simplicity of SGD is both its greatest strength and weakness. Due to the effects of evaluating noisy approximation of the true gradient, SGD achieves a convergence rate which is only sub-linear in the number of steps. In an effort to deal with this randomness, two
primary directions of focus have been developed. The first line of
work focuses on choosing the appropriate SGD step-size~\cite{bach2011, lacoste2012, rakhlin2011}. If a decaying step-size is chosen, the variance is forced to zero asymptotically
guaranteeing convergence. However, small steps also slow down progress
and limit the rate of convergence in practise.  The step-size must be chosen carefully, which can require extensive
experimentation possibly negating the computational speedup of SGD.  Another approach is to use an improved, lower-variance estimate of the gradient. If this estimator is chosen correctly -- such that its variance goes to zero asymptotically -- convergence can be guaranteed with a \emph{constant} learning rate.  This
scheme is used in~\cite{defazio2014, roux2012} where the
improved estimate of the gradient combines stochastic gradients
computed at the current stage with others used at an earlier stage. A
similar approach proposed in~\cite{johnson2013, konevcny2013} combines
stochastic gradients with gradients periodically re-computed at a
pivot point.

With variance reduction, first-order methods can obtain a linear
convergence rate. In contrast, second-order methods have been shown to obtain super-linear convergence. 
However, this requires the computation and inversion of the
Hessian matrix which is impractical for large-scale
datasets. Approximate variants known as quasi-Newton
methods~\cite{dennis1977} have thus been developed, such as the
popular BFGS or its limited memory version known as
LBFGS~\cite{liu1989}. Quasi-Newton methods such as BFGS do not require computing
the Hessian matrix but instead construct a quadratic model of the
objective function by successive measurements of the gradient. This
also yields super-linear convergence when the quadratic model is accurate.
Stochastic variants of BFGS have been proposed (oBFGS \cite{schraudolph2007}), for which stochastic
gradients replace their deterministic counterparts. A
regularized version known as RES~\cite{mokhtari2014b} 
achieves a sublinear convergence rate with a decreasing
step-size by enforcing a
bound on the eigenvalues of the approximate Hessian matrix. SQN~\cite{byrd2014}, another related method also requires a decreasing step size to achieve sub-linear convergence. Although stochastic second order methods
have not be shown to achieve super-linear convergence,
they empirically outperform SGD for problems with a large condition
number~\cite{mokhtari2014b}. 

A clear drawback to stochastic second order methods is that similarly
to their first-order counterparts, they suffer from high variance in
the approximation of the gradient.  Additionally, this problem can be exaggerated due
to the estimate of the Hessian magnifying the effect of this noise. Overall, this can lead to such algorithms taking large steps in poor descent directions.

In this paper, we propose and analyze a stochastic variant of BFGS that uses a
multi-stage scheme similar to~\cite{johnson2013, konevcny2013} to
progressively reduce the variance of the stochastic gradients. We call
this method \underline{V}ar\underline{i}ance-reduced
S\underline{t}ochastic N\underline{e}wton ({\sc Vite}). Under standard conditions on $\Jhat$, we show that that variance reduction
on the gradient estimate alone is sufficient for fast convergence. For smooth and strongly convex functions, {\sc Vite} reaches the optimum at a
geometric rate with a constant step-size. To our knowledge {\sc Vite} is the first stochastic Quasi-Newton method with these properties. 

In the following section, we briefly review the BFGS algorithm and its stochastic variants. We then introduce the {\sc
  VITE} algorithm and analyze its convergence properties. Finally, we present
experimental results on real-world datasets demonstrating its superior performance over a range of competitors.

\section{Stochastic second order optimization}
\subsection{Problem setting}
Given a dataset $\D = \{ (\x_1, y_1), \dots
,(\x_n, y_n )\}$ consisting of feature vectors $\x_i \in \mathbb{R}^d$ and targets $y_i \in [ 0 , C ]$,
we consider the problem of minimizing the expected loss 
$f(\w) = \E [ f_i(\w) ]$. Each
function $f_i(\w)$ takes the form $f_i(\w) = \ell(h(\w, \x_i),
y_i),$ where $\ell$ is a loss function and $h$ is a prediction model
parametrized by $\w \in \mathbb{R}^d$. The expectation is over the set of samples and we denote $\wstar = \arg \min_{\w}
f(\w)$. 

This optimization problem can be solved exactly for convex
functions using gradient descent, where the gradient of the loss
function is expressed as $\dw f(\w) = \E [ \dw f_i(\w) ]$.
When the size of the dataset $n$ is large, the computation of the
gradient is impractical and one has to resort to stochastic
gradients. Similar to gradient descent, stochastic gradient descent
updates the parameter vector $\w_t$ by stepping in the opposite
direction of the stochastic gradient $\dw f_i(\w_t)$ by an amount
specified by a step size $\eta_t$ as follows:
\begin{equation}
\w_{t+1} = \w_t - \eta_t \dw f_i(\w_t).
\label{eq:update_sgd}
\end{equation}

In general, a stochastic gradient can also be computed as an average
over a sample of datapoints as $\hat{f}(\w_t) = r^{-1}\sum_{i=1}^r
f_i(\w_t)$. Given that the stochastic gradients are unbiased estimates
of the gradient, Robbins and Monro~\cite{robbins1951} proved
convergence of SGD to $\wstar$ assuming a decreasing step-size
sequence. A common choice for the step size is
\cite{shalev2011,mokhtari2014b}
\begin{equation}
\it{a) } \; \eta_t = \frac{\eta_0}{t} \quad\quad \text{or} \quad\quad \it{b) } \; \eta_t = \frac{\eta_0 T_0}{T_0 + t} 
\label{eq:step_size}
\end{equation}
where $\eta_0$ is a constant initial step size and $T_0$ controls the speed of decrease.

Although the cost per iteration of SGD is low, it suffers from slow
convergence for certain ill-conditioned
problems~\cite{mokhtari2014b}. An alternative is to use a second order
method such as Newton's method that estimates the curvature of the
objective function and can achieve quadratic
convergence. In the following, we review Newton's method and its
approximations known as quasi-Newton methods.

\subsection{Newton's method and BFGS}

Newton's method is an iterative method that minimizes the Taylor
expansion of $f(\w)$ around $\w_t$:
\begin{align}
f(\w) = & f(\w_t) + ( \w - \w_t)^{\top} \dw f(\w_t) + \frac{1}{2}  ( \w - \w_t)^{\top} H ( \w - \w_t),
\label{eq:taylor_exp}
\end{align}
where $H$ is the Hessian of the function $f(\w)$ and quantifies its curvature. Minimizing
Eq.~\ref{eq:taylor_exp} leads to the following update rule:
\begin{equation}
\w_{t+1} = \w_t - \eta_t H_t^{-1} \cdot \nabla f(\w_t),
\label{eq:newton_update}
\end{equation}
where $\eta_t$ is the step size chosen by backtracking line search.

Given that computing and inverting the Hessian matrix is an expensive
operation, approximate variants of Newton's method have emerged, where
$H_t^{-1}$ is replaced by an approximate version $\tilde{H}_t^{-1}$
selected to be positive definite and as close to $H_t^{-1}$ as
possible. The most popular member of this class of quasi-Newton
methods is BFGS~\cite{nocedal1999} that incrementally updates an
estimate of the inverse Hessian, denoted $J_t =
\tilde{H}_t^{-1}$. This estimate is computed by solving a weighted Frobenius norm minimization subject to the secant
condition:
\begin{equation}
\w_{t+1} - \w_t = J_{t+1} (\nabla f(\w_{t+1}) - \nabla f(\w_t)).
\label{eq:secant}
\end{equation}

The solution can be obtained in closed form leading to the
following explicit expression:
\begin{equation}
J_{t+1} = \left( I - \frac{s y^{\top}}{y^{\top}s} \right) J_t \left( I - \frac{y s^{\top}}{y^{\top}s} \right) + \frac{ss^{\top}}{y^{\top}s},
\label{eq:Sherman-Morrison}
\end{equation}
where $s = \w_{t+1} - \w_t$ and $y = \nabla f(\w_{t+1}) - \nabla f(\w_t)$.
Eq.~\ref{eq:Sherman-Morrison} is known to be positive
definitive assuming that $J_0$ is initialized to be a positive
definite matrix.

\subsection{Stochastic BFGS}

A stochastic version of BFGS (oBFGS) was proposed
in~\cite{schraudolph2007} in which stochastic gradients are used for
both the determination of the descent direction and the approximation
of the inverse Hessian. The oBFGS approach described in
Algorithm~\ref{algo:stochastic_newton} uses the following update
equation:
\begin{equation}
\w_{t+1} = \w_t - \eta_t \Jhat_t \cdot \nabla \hat{f}(\w_t),
\label{eq:oBFGS_update}
\end{equation}
where the matrix $\Jhat_t$ and the vector $\nabla \hat{f}(\w_t)$ are
stochastic estimates computed as follows.
Let $\A \subset \{1 \dots n \}$ and $\B \subset \{1 \dots n
\}$ be sets containing two independent samples of datapoints.  The variables $y$ and
$\nabla f(\w)$ defined in Eq.~\ref{eq:Sherman-Morrison} are replaced
by sampled variables computed as
\begin{equation}
\hat{y} =\frac{1}{|\A|} \sum_{k\in \A} \nabla f_k(\w_{t+1}) - \nabla f_k(\w_t)
\quad \text{and} \quad
\nabla \hat{f}(\w_t) = \nabla f_{\B}(\w_t) = \frac{1}{|\B|}\sum_{k\in\B} \nabla f_k(\w_t).
\label{eq:nabla_f_sampled}
\end{equation}

The estimate of the inverse Hessian then becomes
\begin{equation}
\Jhat_{t+1} = \left( I - \frac{s \yhat^{\top}}{\yhat^{\top} s} \right) \Jhat_t \left( I - \frac{\yhat s^{\top}}{\yhat^{\top} s} \right) + \frac{ss^{\top}}{\yhat^{\top} s}
\label{eq:Sherman-Morrison_2}
\end{equation}

Unlike Newton's method, oBFGS uses a fixed step size sequence instead
of a line search. A common choice is to use a step size similar to the
one used for SGD in Eq.~\ref{eq:step_size}.

\begin{algorithm}[]
\caption{{\bf oBFGS}}
\begin{algorithmic}[1]
\STATE {\bf INPUTS :}
\INDSTATE $\mathcal{D}$ : Training set of $n$ examples.
\INDSTATE $\w_0$ : Arbitrary initial values, e.g., 0.
\INDSTATE $\{\eta_t\}$ : Step size sequence
\STATE {{\bf OUTPUT :} $\w_t$}
\STATE $\Jhat_0 \leftarrow \alpha I$
\FOR{$t=0 \dots T$}
\STATE Randomly pick two sets $\A$ and $\B$
\STATE $s \leftarrow \w_{t+1} - \w_t$
\STATE $\yhat \leftarrow \sum_{k\in\B} \nabla f_k(\w_{t+1}) - \nabla f_k(\w_t)$
\STATE $\nabla \fhat(\w_t) \leftarrow \sum_{k\in \A} \nabla f_k(\w_t)$
\STATE $\w_{t+1} \leftarrow \w_t - \eta_t \Jhat_{t+1} \cdot \nabla \fhat(\w_t)$
\STATE $\Jhat_{t+1} \leftarrow \left( I - \frac{s \yhat^{\top}}{\yhat^{\top} s} \right) \Jhat_t \left( I - \frac{\yhat s^{\top}}{\yhat^{\top} s} \right) + \frac{ss^{\top}}{\yhat^{\top} s}$
\ENDFOR
\end{algorithmic}
\label{algo:stochastic_newton}
\end{algorithm}

A regularized version of oBFGS (RES) was recently proposed
in~\cite{mokhtari2014b}. RES differs from oBFGS in
the use of a regularizer to enforce a bound on the eigenvalues of
$\Jhat_t$ such that
\begin{equation}
\gamma I \preceq \Jhat_t \preceq \rho I = \left( \gamma + \frac{1}{\delta} \right) I,
\label{eq:bound_inv_hessian_res}
\end{equation}
where $\gamma$ and $\delta$ are given positive constants and the notation $A\preceq B$ means that $B - A$ is a positive semi-definite matrix.
Note that \eqref{eq:bound_inv_hessian_res} also implies an upper and lower bound on $\E [ \Jhat_t]$~\cite{mokhtari2014b}.
The update of RES is modified to incorporate an identity bias term
$\gamma I$ as follows:
\begin{equation}
\w_{t+1} = \w_t - \eta_t (\Jhat_t + \gamma I) \cdot \nabla \hat{f}(\w_t).
\label{eq:RES_update}
\end{equation}

The convergence proof derived in~\cite{mokhtari2014b} shows that lower
and upper bounds on the Hessian eigenvalues of the sample functions
are sufficient to guarantee convergence to the optimum. However, the analysis shows that RES will converge to the optimum at a rate $\mathcal{O}(1/t)$ and requires a decreasing step-size. Similar
results were derived in~\cite{byrd2014} for the SQN algorithm.

\section{The {\sc Vite} algorithm}

Reducing the size of the sets $\A$ and $\B$ used to estimate the inverse Hessian approximation and the stochastic gradient is 
desirable for reasons of computational efficiency. However, doing so also increases the variance
of the update step. Here we propose a new method called {\sc Vite}
that explicitly reduces this variance. 

In order to simplify the analysis of {\sc Vite}, we do not explicitly consider the randomness in the matrix $\Jhat_t$. Instead, we assume that it is positive definite (which holds under weak conditions due to the BFGS update step) and that its variance can be kept under control, for example by using the regularization of the RES method.

To motivate {\sc Vite} we first consider the standard oLBFGS step, \eqref{eq:oBFGS_update} estimated with the sets $\A$ and $\B$. The first and second moments simplify as
\begin{align}
\E ~ [ \Jhat_t  \nabla f_{\B}(\w_t)] = \Jhat_t  \E_{\B} [ \nabla f_{\B}(\w_t) ]
\label{eq:E_Newton_direction}
\end{align}
and
\begin{align}
 \E ~ & \sqnorm{ \Jhat_t  \nabla f_{\B}(\w_t)} 
\leq \sqnorm{\Jhat_t} \E_{\B} \sqnorm{ \nabla f_{\B}(\w_t)},
\label{eq:bound_second_moment}
\end{align}
respectively.
\comment{
In order to reduce the variance of the estimate $\Jhat_t \cdot
\nabla f_{\B}(\w_t)$, one can thus reduce the variance of $\Jhat_t$
and the variance of $\nabla \hat{f}(\w_t)$ independently. Here we assume
that the variance of $\Jhat_t$ can be kept under control, for example
using the regularization of the RES method. On the other hand, we
propose to reduce the variance of $\nabla \hat{f}(\w_t)$ using a variance
reduction technique similar to the one proposed in~\cite{johnson2013,
  konevcny2013}.
  }
For $|\A|$ large enough, in order to reduce the variance of the estimate $\Jhat_t \cdot
\nabla f_{\B}(\w_t)$, it is only required to reduce the variance of $\nabla {f}_{\B}(\w_t)$ independently. We
proceed using a technique similar to the one proposed in~\cite{johnson2013,
  konevcny2013}.

{\sc Vite} differs from oBFGS and other stochastic Quasi-Newton methods in the
use of a multi-stage scheme as shown in Algorithm~\ref{algo:hvrg}. In the outer loop a variable $\wtil$ is introduced. We periodically evaluate the gradient of the function with respect to $\wtil$.
This \emph{pivot point} is inserted in the update
equation to reduce the variance. Each inner loop runs for a
a random number of steps $t_j \in [1, m]$ whose distribution follows a
geometric law with parameter
$\beta = \sum_{t=1}^m (1 - \mu \gamma \eta)^{m-t}$.
Stochastic gradients at $\w_t$ and $\wtil$ are computed and the inverse Hessian approximation is updated
in each iteration of the inner loop. $\Jhat_t$ can be updated using the same update as RES although we found in practice
that using Eq.~\ref{eq:Sherman-Morrison_2} did not
affect the results significantly. 
  The descent direction $\nabla
f_{\B}(\w)$ is then replaced by 
$$\vt = \nabla f_{\B}(\w_t) - \nabla
f_{\B}(\wtil) + \mutil.$$ 
{\sc Vite} then makes updates of the form
\begin{equation}
\w_{t+1} = \w_t - \eta \Jhat_t \cdot \vt.
\label{eq:VITE_update}
\end{equation}
Clearly, $\mutil = \E [\nabla
  f_{\B}(\wtil)]$ and $\E [\vt] = \E [\nabla f_{\B}(\w_t)]$ so in expectation the descent is in the same direction as Eq.~\eqref{eq:E_Newton_direction}. Following the analysis of \cite{johnson2013}, the variance of $\vt$ goes to zero when both $\wtil$ and $\w_t$ converge to the
same parameter $\wstar$. Therefore, convergence can be guaranteed with a \emph{constant}
step-size. The complexity of this approach depends on the number of epochs $S$
and a constant $m$ limiting the number of stochastic gradients
computed in a single epoch, as well as other parameters that will be
introduced in more detail in Section~\ref{sec:analysis}.

\begin{algorithm}[]
\caption{{\sc Vite}}
\begin{algorithmic}[1]
\STATE {\bf INPUTS :}
\INDSTATE $\mathcal{D}$ : Training set of $n$ examples \quad\quad\quad $\wtil_0$ : Arbitrary initial values, e.g., 0
\INDSTATE $\eta$ : Constant step size \quad\quad\quad\quad\quad\quad\quad $m$: Arbitrary constant
\STATE {{\bf OUTPUT :} $\w_t$}
\STATE $\Jhat_0 \leftarrow \alpha I$
\FOR{$s=0 \dots S$}
\STATE $\wtil = \wtil_{s-1}$
\STATE $\mutil = \frac{1}{n} \sum_{i=1}^n \nabla f_i(\wtil)$
\STATE $\w_0 = \wtil$
\STATE Let $t_j \leftarrow t$ with probability $\frac{(1 - \mu \rho \eta)^{m-t}}{\beta}$ for $t = 1, \dots, m$
\FOR{$t=0 \dots t_j - 1$}
\STATE Randomly pick independent sets $\A, \B \subset \{1 \dots n\}$
\STATE $\vt = \nabla f_{\B}(\w_t) - \nabla f_{\B}(\wtil) + \mutil$
\STATE $\w_{t+1} \leftarrow \w_t - \eta \Jhat_{t} \cdot \vt$
\STATE Update $\Jhat_{t+1}$
\ENDFOR
\STATE $\wtil_{s} = \w_{t_j}$.
\ENDFOR
\end{algorithmic}
\label{algo:hvrg}
\end{algorithm}

\newcommand{\state}{{\mathbf w}}
\newcommand{\stateb}{{\mathbf v}}
\newcommand{\mineigen}{{\lambda_{\text{min}}}}
\newcommand{\maxeigen}{{\lambda_{\text{max}}}}
\newcommand{\Jbar}{{\bar J}}
\section{Analysis}
\label{sec:analysis}

In this section we present a convergence proof for the {\sc Vite} algorithm that builds upon and generalizes previous analyses of variance reduced first order methods \cite{johnson2013, konevcny2013}. Specifically, we show how variance reduction on the stochastic gradient direction is sufficient to establish geometric convergence rates, even when performing linear transformations with a matrix $\Jhat_{t}$. Since we do not exploit the specific form of the stochastic evolution equations for $\Jhat_t$, this analysis will not allow us to argue in favor of the specific choice of Eq.~\eqref{eq:Sherman-Morrison_2}, yet it shows that variance reduction on the gradient estimate is sufficient for fast convergence as long as $\Jhat_t$ is sufficiently well behaved.
Our analysis relies on the following standard assumptions:

\paragraph{A1} Each function $f_i$ is differentiable and has a Lipschitz continuous gradient with constant $L>0$, i.e. $\forall \state,\stateb \in \mathbb{R}^n$,
\begin{equation}
f_i(\state) \leq f_i(\stateb) + (\state-\stateb)^{\top} \nabla f_i(\stateb) + \frac{L}{2}
\sqnorm{\state - \stateb}
\label{eq:lipschitz}
\end{equation}

\paragraph{A2} $f$ is $\mu$-strongly convex, i.e. $\forall \state,\stateb \in \mathbb{R}^n$,
\begin{equation}
f(\state) \geq f(\stateb) + (\state-\stateb)^{\top} \nabla f(\stateb) + \frac{\mu}{2} \sqnorm{\state-\stateb}
\label{eq:strongly_convex}
\end{equation}
which also implies
\begin{equation}
\sqnorm{\nabla f(\w)} \geq 2 \mu ( f(\w) - f(\wstar) ) \;\; \forall \w \in \mathbb{R}^n
\label{eq:strongly_convex_2}
\end{equation}
for the minimizer $\wstar$ of $f$. 

Assumptions {\bf A1} and {\bf A2} also implies that the eigenvalues of the Hessian are bounded as follows:
\begin{equation}
\mu I \preceq H_t \preceq  L I .
\label{eq:bound_true_hessian}
\end{equation}
Finally we make the assumption that the inverse Hessian approximation is always well-behaved.
\paragraph{A3} There exist positive constants $\gamma$ and $\rho$ such that, $\forall \w \in \mathbb{R}^n$,
\begin{equation}
\gamma I \preceq \Jhat_t \preceq \rho I .
\label{eq:bound_inv_hessian}
\end{equation}
Assumption {\bf A3} is equivalent to assuming that $\Jhat_t$ is bounded in expectation (see: e.g.~\cite{mokhtari2014b}) but allows us to remove this complication, simplifying notation in the analysis which follows.
We now introduce two lemmas required for the proof of convergence.
\begin{mylemma}
\label{lemma:weighting}
The following identity holds:
\begin{align*}
\E f(\wtil_{s+1}) = \frac{1}{\beta} \sum_{t=0}^{m-1} \tau_t \E f(\w_{t})
\end{align*}
where $ \tau_t := (1 - \gamma \eta \mu)^{m-t-1}$ and the weight vectors $\w_t$ belong to epoch $s$.
\end{mylemma}
This result follows directly from Lemma 3 in ~\cite{konevcny2013}.
\begin{mylemma}
\begin{align}
\E \| \vt \|^2  &\leq 4L (f(\w_{t}) - f(\wstar) + f(\wtil) - f(\wstar)) \nonumber
\end{align}
\end{mylemma}
The proof is given in \cite{johnson2013, konevcny2013} and reproduced for convenience in the Appendix. We are now ready to state our main result.

\begin{mytheorem}
Let Assumptions {\bf A1}-{\bf A3} be satisfied. Define the rescaled strong convexity $\mu' := \gamma\mu \leq \mu$ and Lipschitz $ L' := \rho L \geq L$ constants respectively.  Choose $0 < \eta \leq \frac{\mu'}{2 {L'}^2}$ 
and let m be sufficiently large so that
$\alpha = \frac{(1- \eta\mu')^m}{\beta \eta ( \mu' - 2 {L'}^2\eta )} + \frac{ 2{L'}^2 \eta }{   \mu' - 2{L'}^2 \eta } < 1.$

Then the suboptimality of $\wtil_s$ is bounded in expectation as follows:
\begin{align}
\E (f(\wtil_{s}) - f(\wstar) &\leq \alpha^s \E [f(\w_0) - f(\wstar) ].
\end{align}
\end{mytheorem}
\begin{myrem}
Observe that $\gamma$ and $\rho$ are bounds on the \emph{inverse} Hessian approximation. If $\Jhat_t$ is a good approximation to $H$, then by plugging in $\gamma = L$ and $\rho = \mu$, the upper bound on the learning rate reduces to $\eta \leq \frac{1}{2 \mu L}$.
\end{myrem}
\begin{proof}[Proof of Theorem 1.]
Our starting point is the basic inequality
\begin{align}
f(\w_{t+1}) & = f(\w_t - \eta \Jhat_{t} \cdot \vt) \nonumber \\
&\leq f(\w_{t}) - \eta \langle \nabla f(\w_t), \Jhat_{t} \cdot \vt \rangle + \frac{L}{2} \eta^2 \sqnorm{\Jhat_t \vt} .
\label{eq:basic_eq}
\end{align}
We first use the properties of $\vt$ and $\Jhat_t$ to reduce the dependence of \eqref{eq:basic_eq} on $\Jhat_t$ to its largest and smallest eigenvalues given by \eqref{eq:bound_inv_hessian}. For the purpose of the analysis, we define $\F_{t}$ to be the sigma-algebra measuring $\w_{t}$. By conditioning on $\F_{t}$, and by {\bf A3}, the remaining randomness is in the choice of the index set $\B$ in round $t$, which is tied to the stochasticity of $\vt$. Taking expectations with respect to $\B$ gives us 
\begin{align}
\E_{\B} & \sqnorm{\Jhat_t  \vt}\leq  \| \Jhat_t\|^2 \E_{\B} \| \vt \|^2 \leq \rho^2 \E_{\B} \| \vt \|^2 
\end{align}
and
\begin{align} \label{eq:E_b}
\E_{\B} \langle  \nabla f(\w_t), \Jhat_{t} \cdot \vt \rangle = \langle  \nabla f(\w_t), \Jhat_{t} \cdot \nabla f(\w_t)\rangle 
\geq \gamma \sqnorm{\nabla f(\w_t)}
\end{align}
where  \eqref{eq:E_b} comes from the definition $\E_{\B} \vt = \nabla f(\w_t)$.
Therefore, taking the expectation of the inequality~\eqref{eq:basic_eq} and dropping the notational dependence on $\B$ results in
\begin{align}
\E f(\w_{t+1}) & 
\leq \E  f(\w_{t}) - \gamma \eta \E \sqnorm{\nabla f(\w_{t})} + \frac{L}{2}\eta^2 \rho^2 \E \sqnorm{\vt} .
\label{eq:exp_inequality}
\end{align}

\comment{
\begin{align}
\E[f(\w_{t+1})] & \leq \E[f(\w_{t})] 
+ 4 \frac{\Lambda}{2} \eta^2 \rho^2 L \E \big( f(\w_{t}) - f(\wstar) + f(\wtil_s) - f(\wstar)\big) \\
& - 2 \gamma \eta \mu \E \big( f(\w_t) - f(\wstar) \big) \nonumber \\
&= \E[f(\w_{t})] 
+ 2 \eta^2 \rho^2 \Lambda L \big(f(\wtil_s) - f(\wstar)\big)
+ 2\big(  \eta^2 \rho^2 \Lambda L - \gamma \eta \mu \big) \E [ f(\w_{t}) - f(\wstar) ] \nonumber
\end{align}
}

To simplify the remainder of the proof we make the following substitution
$$
\mu' := \gamma\mu \leq \mu \quad \text{and} \quad L' := \rho L \geq L .
$$

Considering a fixed epoch $s$, we can further bound $\E f(\w_{t+1})$ using
Lemma 2 and Eq.~\ref{eq:strongly_convex_2}. By taking the expectation
over $\F_{t}$, adding and subtracting $f(\wstar)$, we get
\begin{align}
\E[f(\w_{t+1}) - f(\wstar)]  \leq & \E[f(\w_{t}) - f(\wstar)]
+ 2 \eta^2 {L'}^2 \big(f(\wtil_s) - f(\wstar)\big)  \\
& + 2\big(  \eta^2 {L'}^2 - \eta \mu' \big)  \E [ f(\w_{t}) - f(\wstar) ] \nonumber \\
 = &  2 \eta^2 {L'}^2 \big(f(\wtil_s) - f(\wstar)\big)
+ \big( 2 \eta^2 {L'}^2 - 2 \eta \mu'  + 1 \big) \E [ f(\w_{t}) - f(\wstar) ] . \nonumber
\end{align}

Writing $\Delta f(\w_t) := f(\w_t) - f(\wstar)$,
we then have
\begin{align}
(\eta\mu' - 2\eta^2{L'}^2) \E \Delta f(\w_{t}) & \leq 2 \eta^2  {L'}^2  \Delta f(\wtil_{s}) + (1 - \eta\mu' ) \E \Delta f(\w_{t}) - \E \Delta f(\w_{t+1})
\end{align}

Now we sum all these inequalities at iterations $t=0,\dots,m-1$
performed in epoch $s$ with weights $\tau_t = (1 - \eta\mu')^{m-t-1}$. Applying
Lemma \ref{lemma:weighting} to the last summand to recover
$f(\wtil_{s+1})$ we arrive at
\begin{align}
\beta \E \Delta f(\wtil_{s+1}) & \leq \frac{2\beta \eta^2{L'}^2}{\eta\mu'-2\eta^2{L'}^2}  \E \Delta f(\wtil_{s}) + \sum_{t=0}^{m-1} \tau_t \frac{(1- \eta \mu') \E \Delta f(\w_{t}) - \E \Delta f(\w_{t+1})}{\eta\mu'-2\eta^2 {L'}^2} .
\nonumber
\end{align}

We now need to bound the remaining sum $(*)$ in the numerator, which can be accomplished by re-grouping summands
\begin{align}
(*) = & (1 - \eta\mu')^{m} \E \triangle f(\wtil_{s}) - (1 - \eta\mu') \E \triangle f(\wtil_{s+1})
\nonumber
\end{align}

By ignoring the negative term in $(*)$, we get the final bound
\begin{align}
\E \Delta f(\wtil_{s+1}) & \leq \alpha \E \Delta f(\wtil_{s}),
\nonumber
\end{align}
where
$$
\alpha = \left( \frac{(1-\eta\mu')^m}{\beta(\eta\mu'-2\eta^2 {L'}^2)} + \frac{ 2\eta^2 {L'}^2}{\eta\mu'-2\eta^2{L'}^2} \right)
$$
\end{proof}

Theorem 1 implies that {\sc Vite} has a local geometric convergence rate
with a constant learning rate. In order to satisfy $\E (f(\wtil_{s}) -
f(\wstar)) \leq \epsilon$, the number of stages $s$ needs to satisfy
$$s \geq - \log \alpha^{-1} \log \frac{\E (f(\wtil_{0}) -
  f(\wstar))}{\epsilon}.$$

Since each stage requires $n + m (2|\A| +
2|\B|)$ component gradient evaluations, the overall complexity is
$
\mathcal{O}((n + 2m (|\A| + |\B|)) \log(1/\epsilon)).
$

\section{Experimental Results}

\begin{figure*}[t!]
	\begin{center}
          \begin{tabular}{@{}W@{\hspace{1mm}}T@{\hspace{1mm}}T@{\hspace{1mm}}T@{}}
            {\large {$|\B| = 1$}} &
            \includegraphics[trim = 15 15 15 15, clip, width=0.8\linewidth]{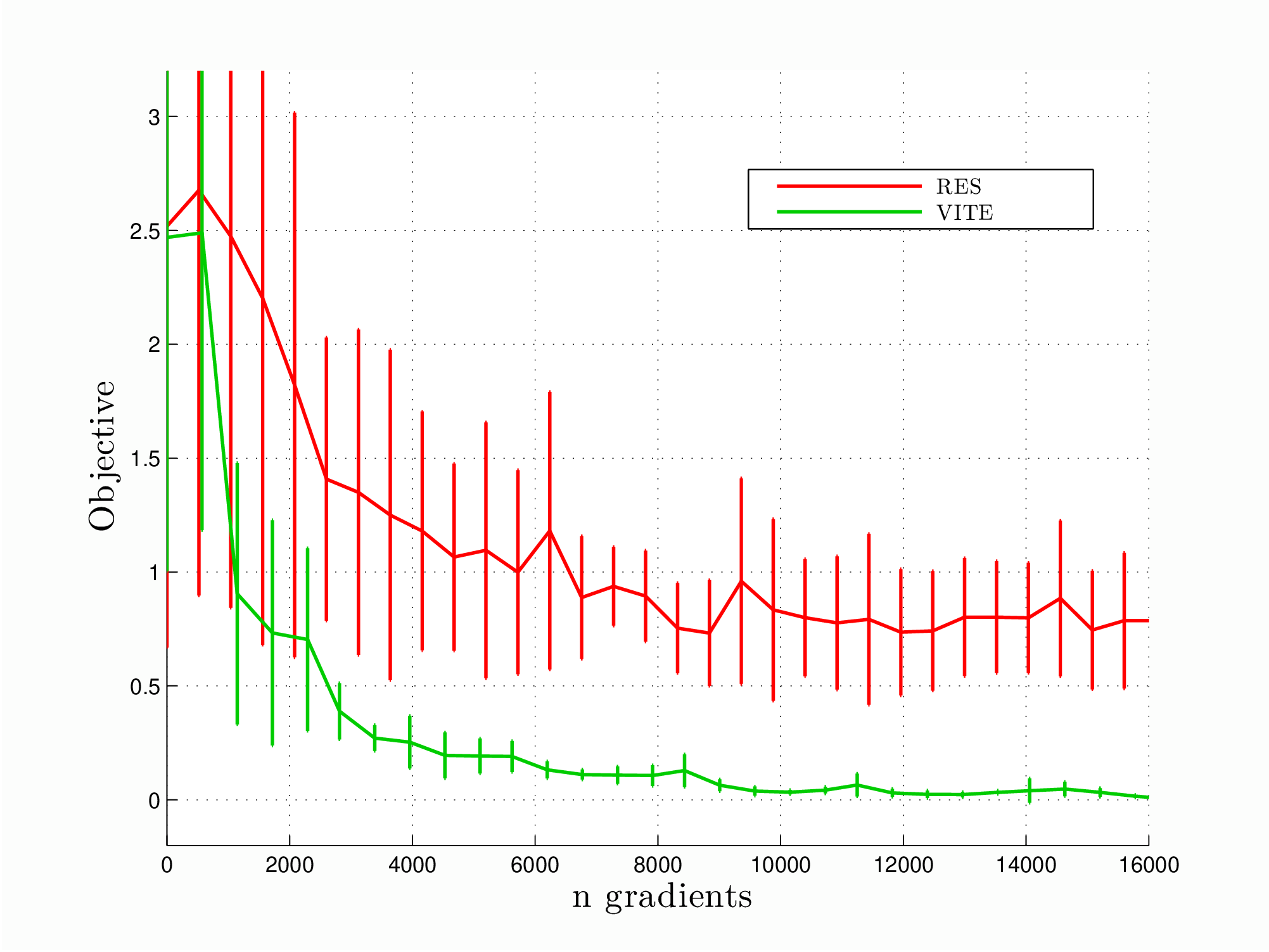} &
            \includegraphics[trim = 15 15 15 15, clip, width=0.8\linewidth]{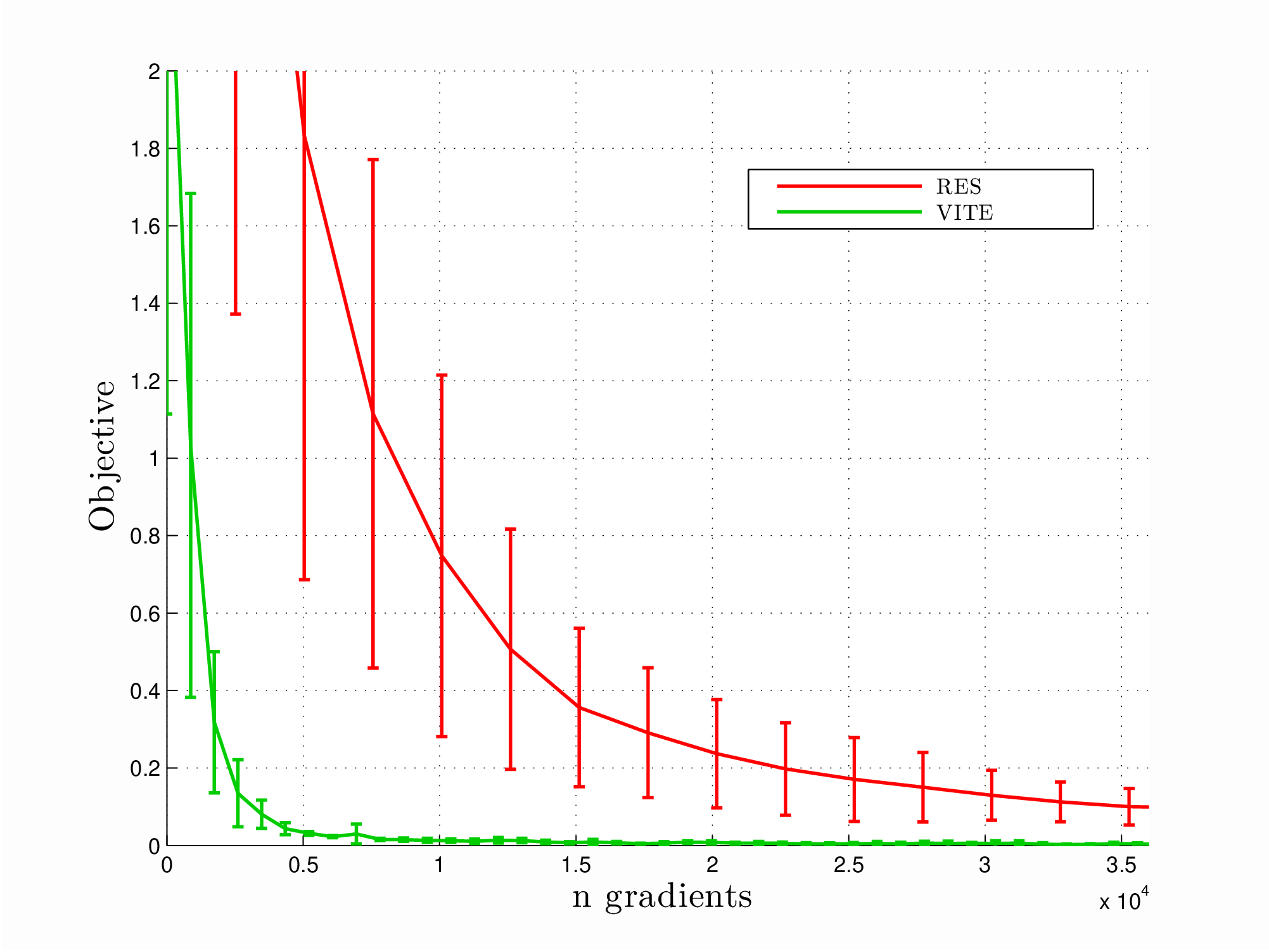} &
            \includegraphics[trim = 15 15 15 15, clip, width=0.8\linewidth]{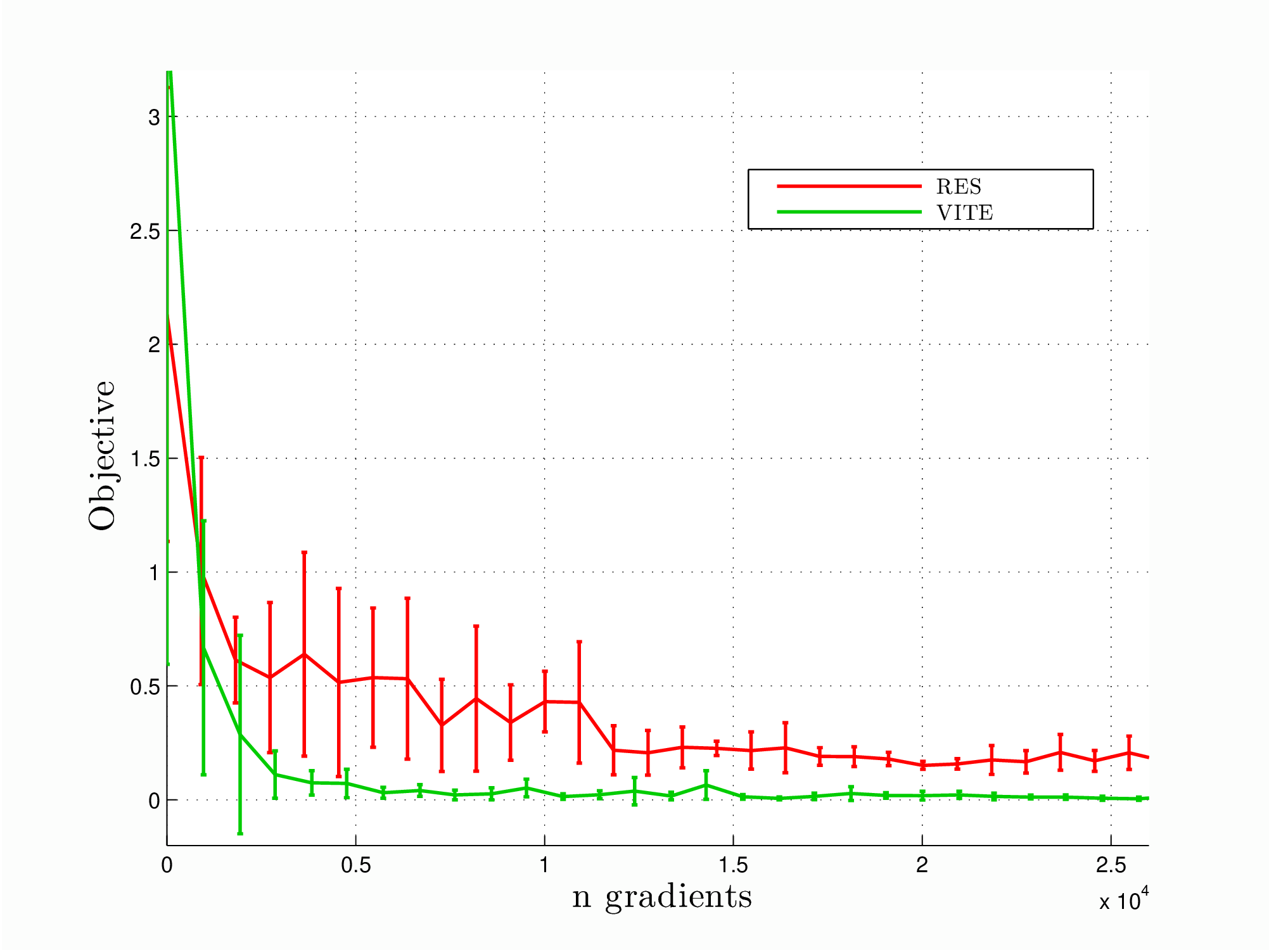} \\
            {\large {$|\B| = 0.1\%$}} &
            \includegraphics[trim = 15 15 15 15, clip, width=0.8\linewidth]{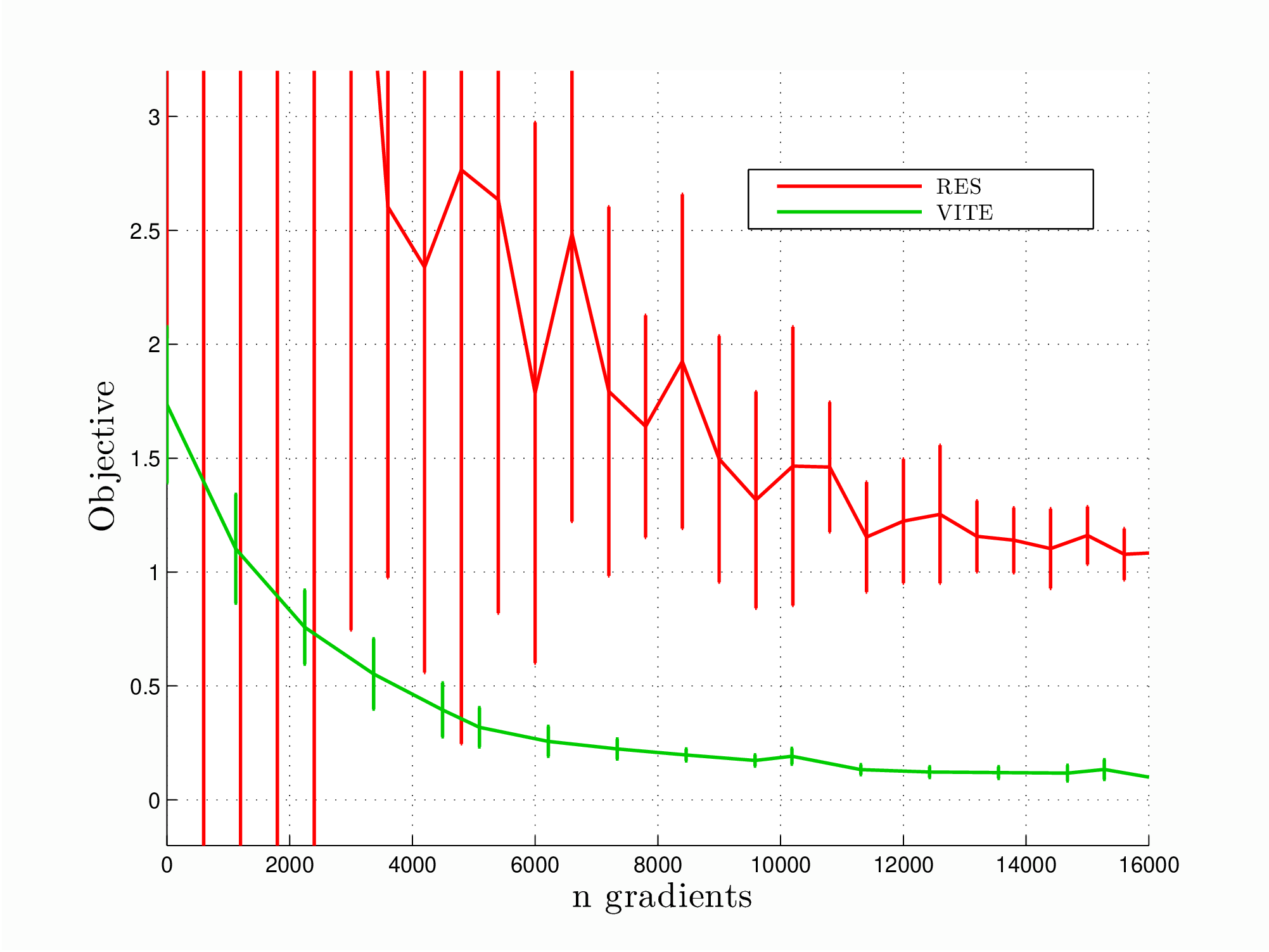} &
            \includegraphics[trim = 15 15 15 15, clip, width=0.8\linewidth]{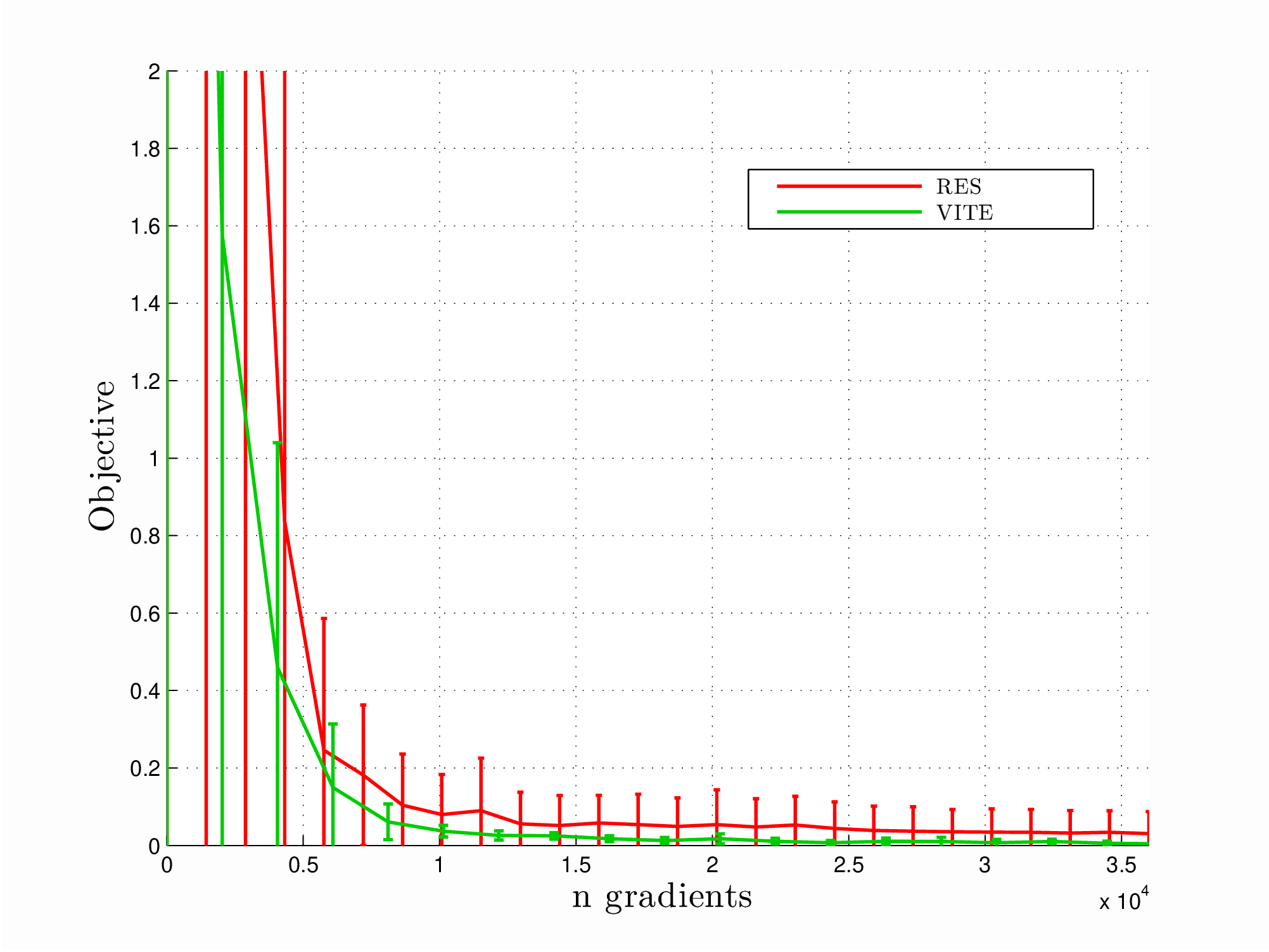} &
            \includegraphics[trim = 15 15 15 15, clip, width=0.8\linewidth]{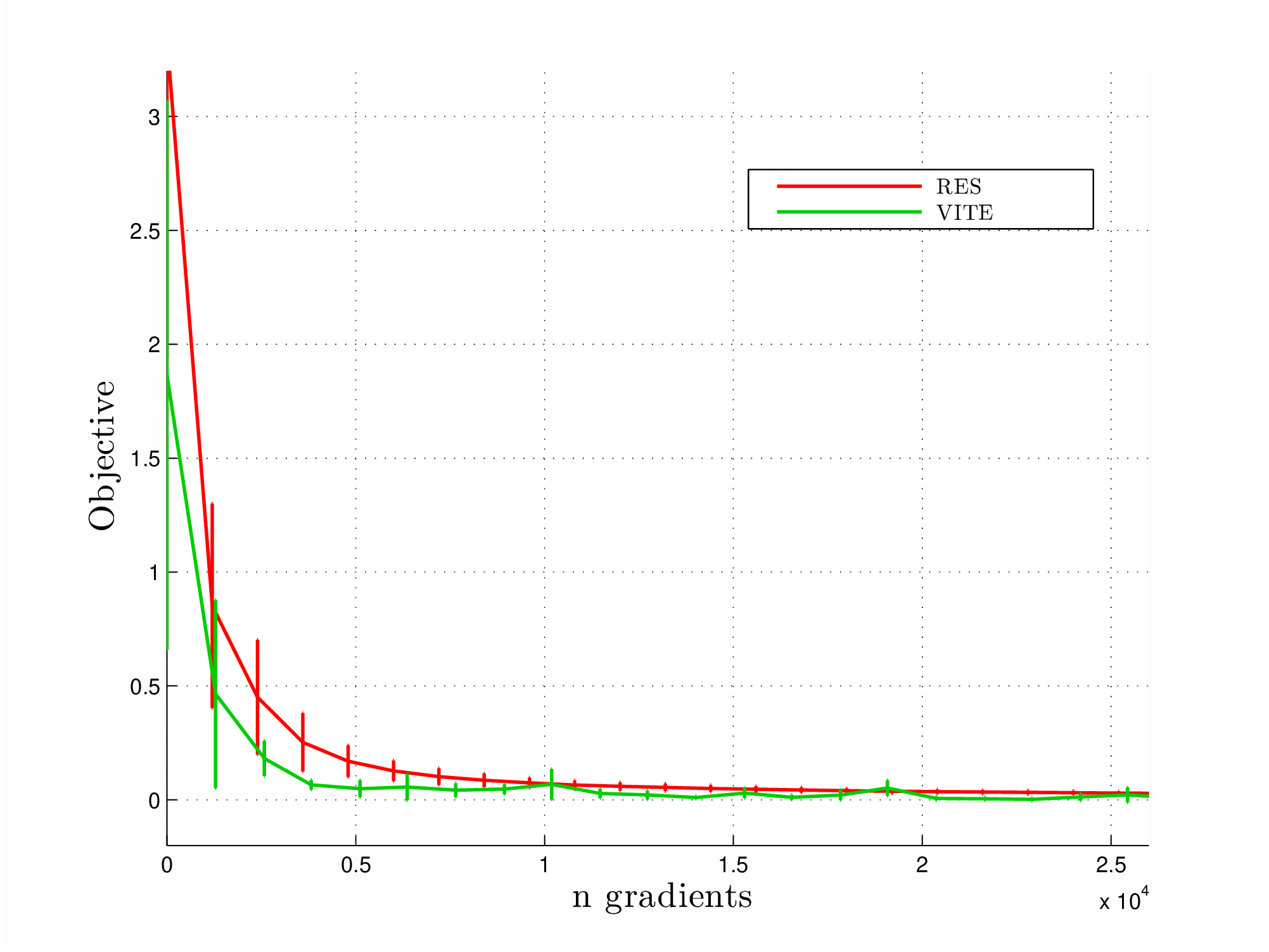} \\
            {\large {$|\B| = 2.5\%$}} &
            \includegraphics[trim = 15 15 15 15, clip, width=0.8\linewidth]{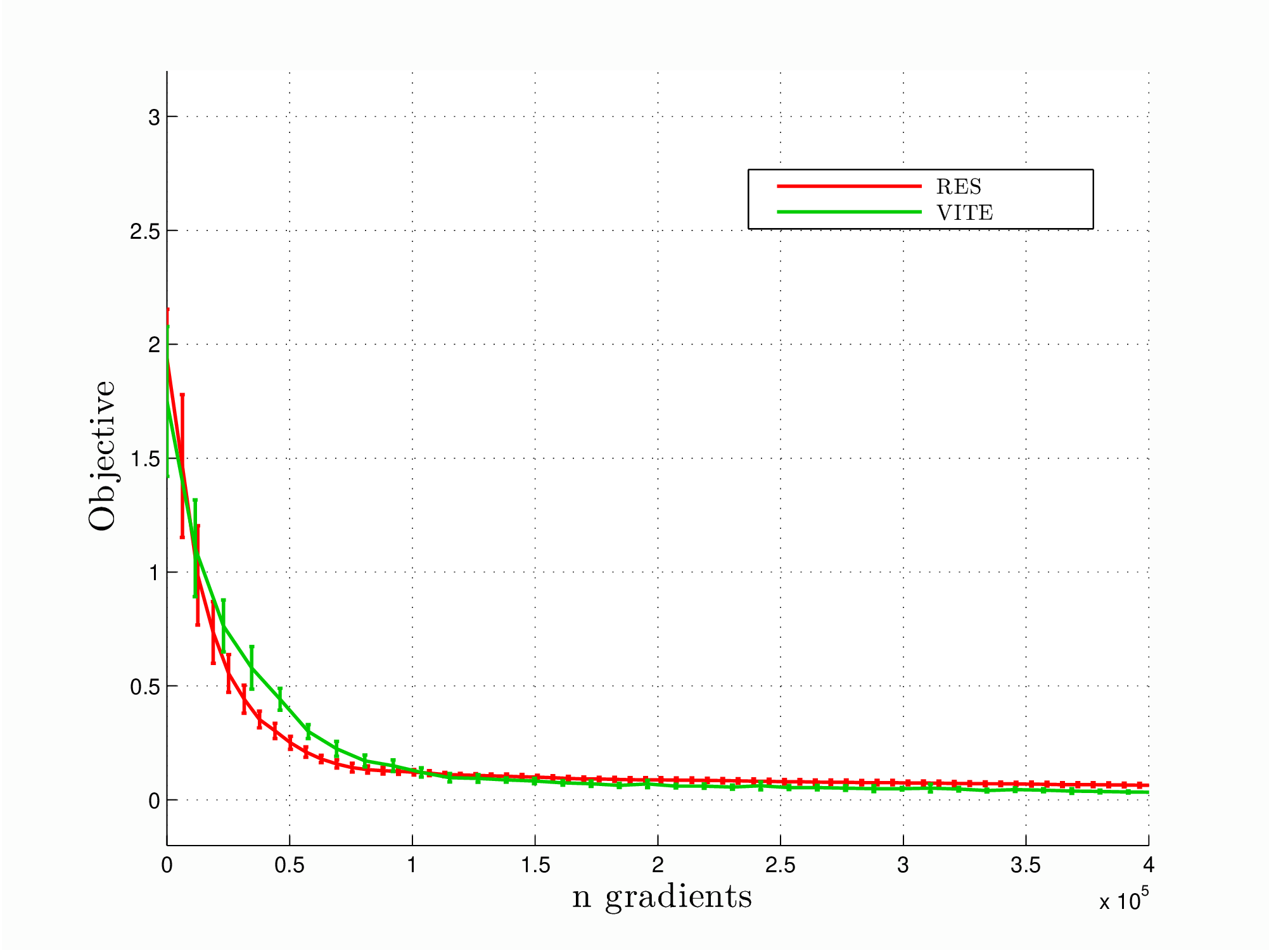} &
            \includegraphics[trim = 15 15 15 15, clip, width=0.8\linewidth]{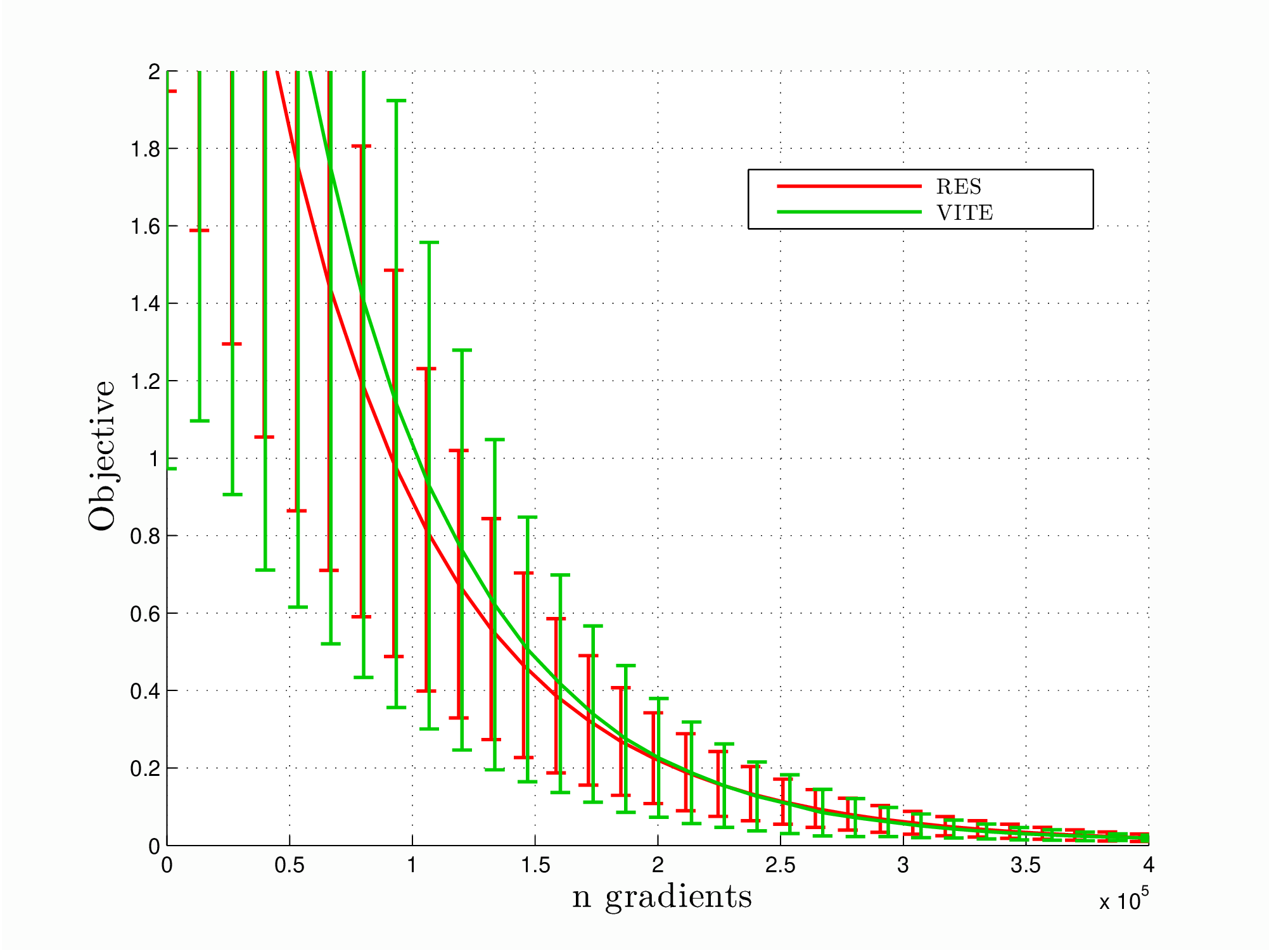} &
            \includegraphics[trim = 15 15 15 15, clip, width=0.8\linewidth]{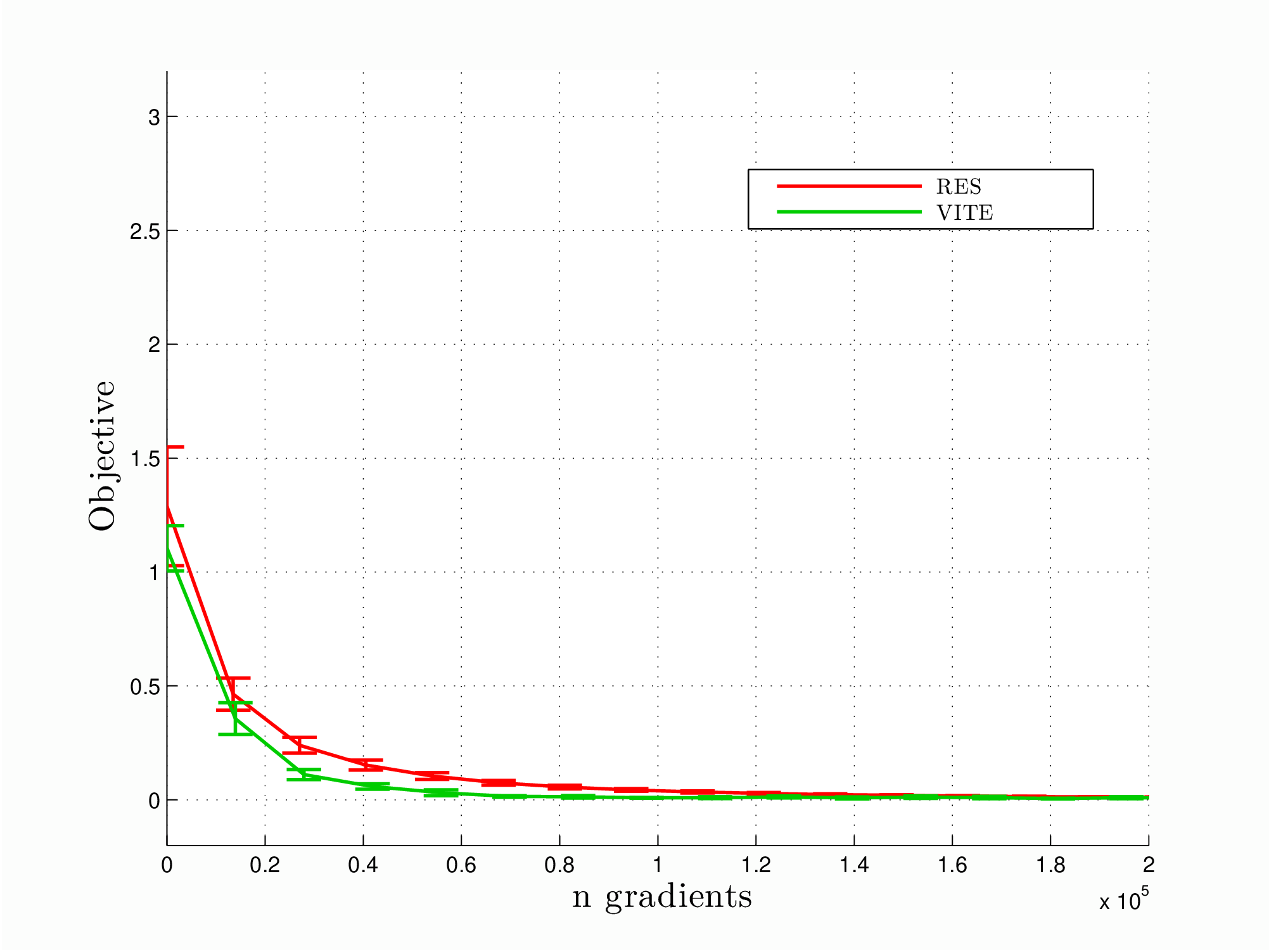} \\
            &
            (a) {\sc Cov} &
            (b) {\sc Adult} &
            (c) {\sc Ijcnn}
            \\
	  \end{tabular}
          \caption{\footnotesize{{\it The red and green curves are the losses achieved
            by RES and {\sc Vite} respectively for varying size of $|\B|$ as a percentage of $n$. Each experiment was
            averaged over 5 runs. Error bars denote variance. In the regime $|\B| \leq 0.1\%$,
            {\sc Vite} has a much lower variance and reaches a lower
            optimum value. Increasing $|\B|$ further decreases the variance of the stochastic gradients but requires more gradient evaluations, decreasing the gap in performance between the methods. Overall, we found {\sc
              VITE} with $|\B|=1\%$ and $|\B|=0.1\%$ to perform the best.}}}
          \label{fig:results}
	\end{center}
\end{figure*}

This section presents experimental results that compare the performance of {\sc Vite} to SGD, SVRG \cite{johnson2013} which incorporates variance reduction and RES \cite{mokhtari2014b} which incorporates second order information. We consider
two commonly occurring problems in machine learning, namely
least-square regression and regularized logistic regression.

{\bf Linear Least Squares Regression.}
We apply least-square regression on the binary version of the
{\sc Cov} dataset~\cite{collobert2002} that contains
$n = 581,012$ datapoints, each described by $d=54$ input features.
{\bf Logistic Regression.}
We apply logistic regression on the {\sc Adult} and {\sc Ijcnn1}
datasets obtained from the {\sc LibSVM} website
\footnote{\url{http://www.csie.ntu.edu.tw/~cjlin/libsvmtools/datasets}}. The
         {\sc Adult} dataset contains $n = 32,561$ datapoints, each
         described by $d = 123$ input features. The {\sc Ijcnn1}
         dataset contains $n = 49,990$ datapoints, each described by
         $d = 22$ input features. We added an $\ell_2$-regularizer with parameter $\lambda=10^{-5}$ to ensure
the objective is strongly convex.

The complexity of {\sc Vite} depends on three quantities: the
approximate Hessian $\Jhat$, the pair of stochastic gradients $(\nabla
f_{\B}(\w), \nabla f_{\B}(\wtil))$ and $\mutil$, respectively computed
over the sets $\A$, $\B$ and $\D$.  Similarly to~\cite{mokhtari2014b},
we consider different choices for $|\A|$ and $|\B|$ and pick
the best value in a limited interval $\{ 1 , \dots, 0.05 n\}$. These results are also reported for the RES method that also
depends on both $|\A|$ and $|\B|$. For SGD, we use $|\B| = 1$ as we
found this value to be the best performer on all datasets. Computing the average gradient, $\mutil$ over the full dataset for SVRG and
{\sc Vite} is impractical. We therefore estimate 
$\mutil$ over a small subset $\C \subset \D$. Although this introduces
some bias, it did not seem to practically
affect convergence for sufficiently large $|\C|$. In our experiments, we
 selected $|\C|=0.1 n$ samples uniformly at random. Each
experiment was averaged over 5 runs with different initializations of
$\w_0$ and a random selection of the samples in $\A$, $\B$
and $\C$. Given that the complexity per iteration of each method is
different, we compare them as a function of the number of gradient
evaluations.

Fig.~\ref{fig:results} shows the empirical convergence properties of
{\sc Vite} against RES for least-square regression and logistic
regression. The horizontal axis corresponds to the number of gradient
evaluations while the vertical axis corresponds to the objective
function value. The vertical bars in each plot show the variance over
5 runs. We show plots for different values of $|\B|$ and the best
corresponding $\A$. For small $|\B|$, the variance of the stochastic
gradients clearly hurts RES while the variance corrections of {\sc
  Vite} lead to fast convergence. As we increase $|\B|$, thus reducing
the variance of the stochastic gradients, the convergence rate of RES
and {\sc Vite} becomes similar. However, {\sc Vite} with small $|\B|$
is much faster to converge to a lower objective value. This clearly
demonstrates how using small batches for the computation of the
gradients while reducing their variance leads to a fast convergence
rate.  We also investigated the effect of $|\A|$ on the convergence of
RES and {\sc Vite} (see Appendix). In short, we find that a good-enough curvature estimate can be obtained for $|\A|=\mathcal{O}(10^{-5}n)$. Increasing this value incurs a penalty in terms of number of gradient evaluations required and so overall performance degrades.

\begin{figure*}[t!]
	\begin{center}
          \begin{tabular}{@{}c@{\hspace{1mm}}c@{\hspace{1mm}}c@{}}
            \includegraphics[trim = 15 15 15 15, clip, width=0.29\linewidth]{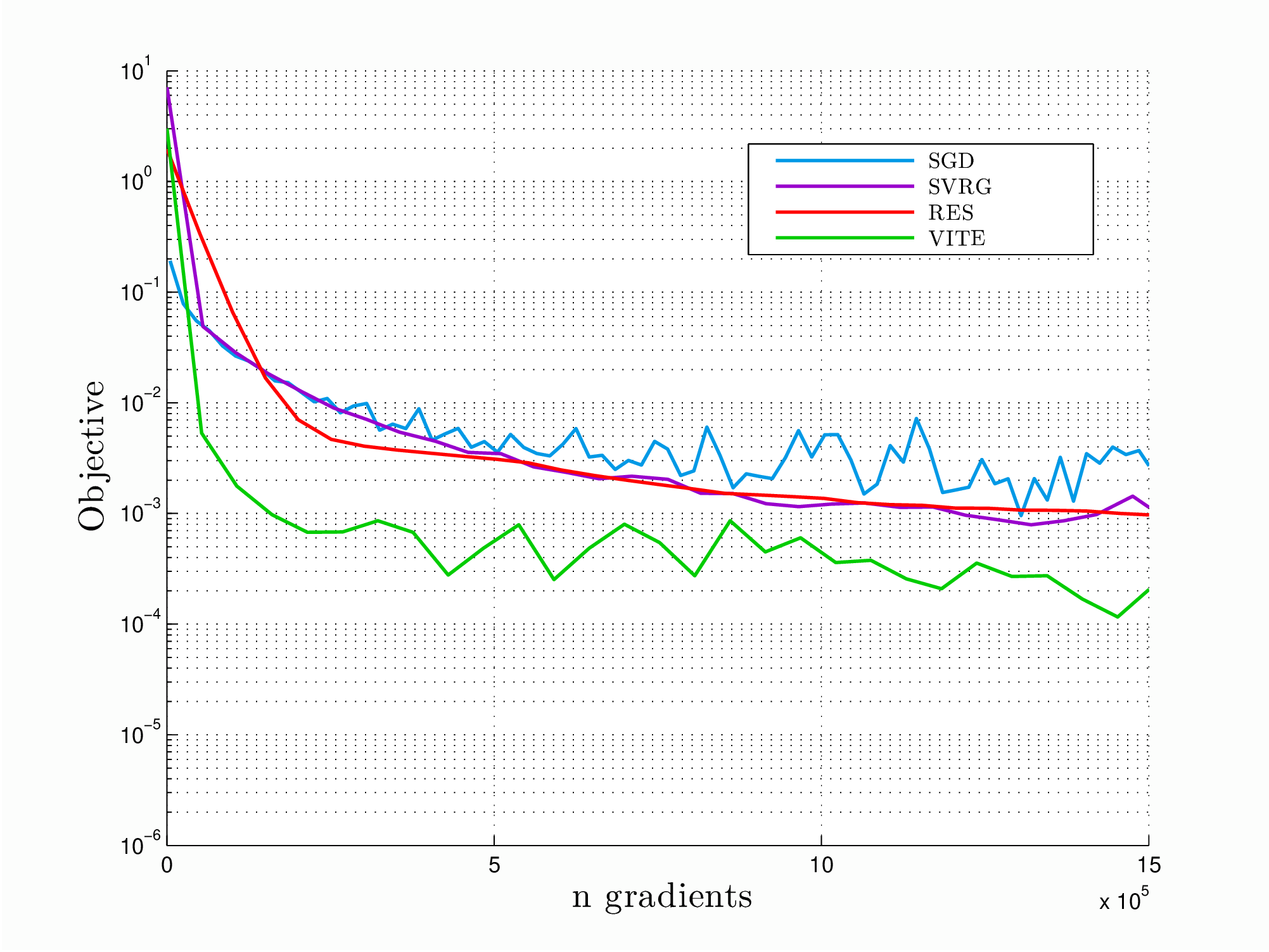} &
            \includegraphics[trim = 15 15 15 15, clip, width=0.29\linewidth]{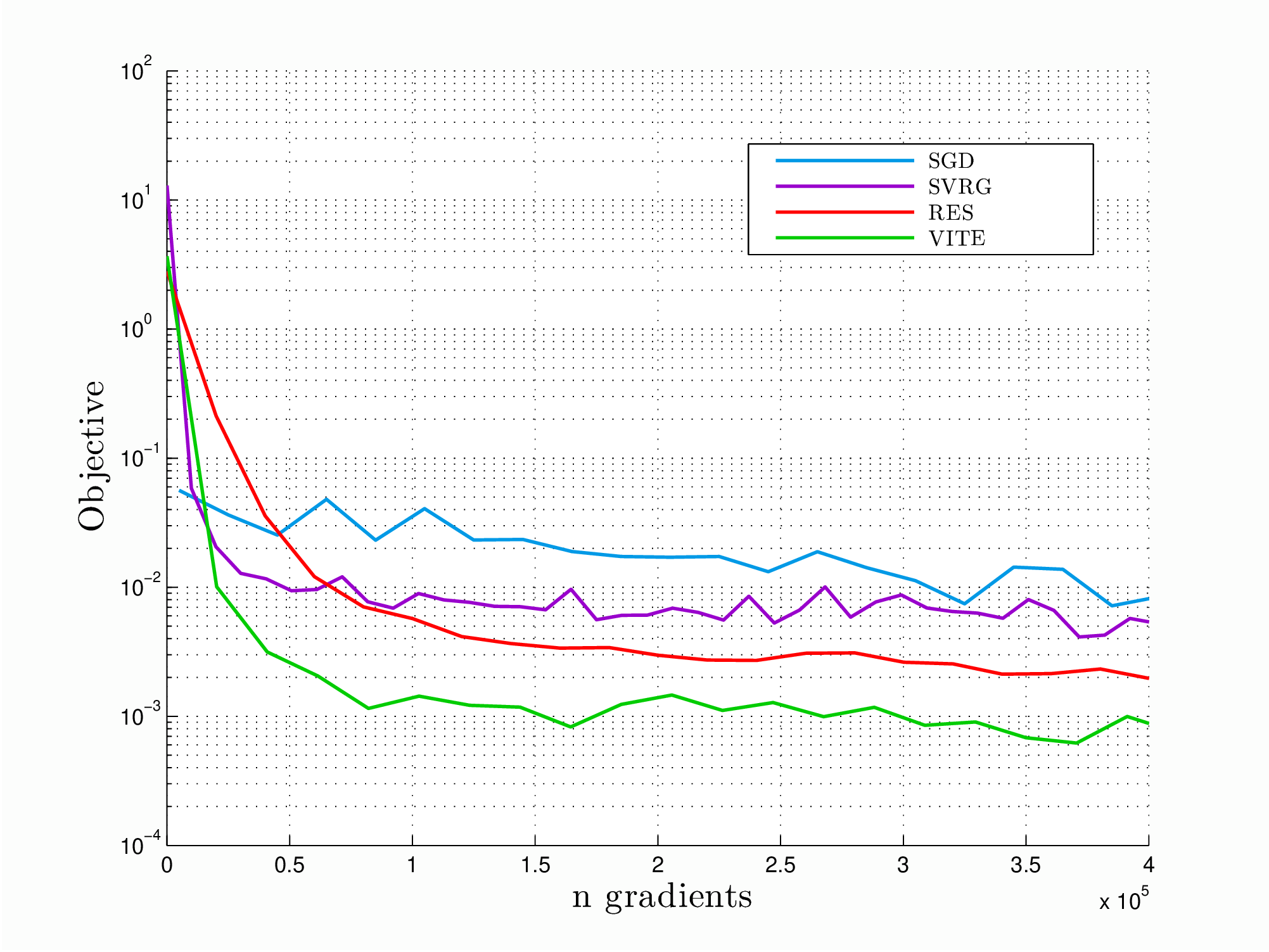} &
            \includegraphics[trim = 15 15 15 15, clip, width=0.29\linewidth]{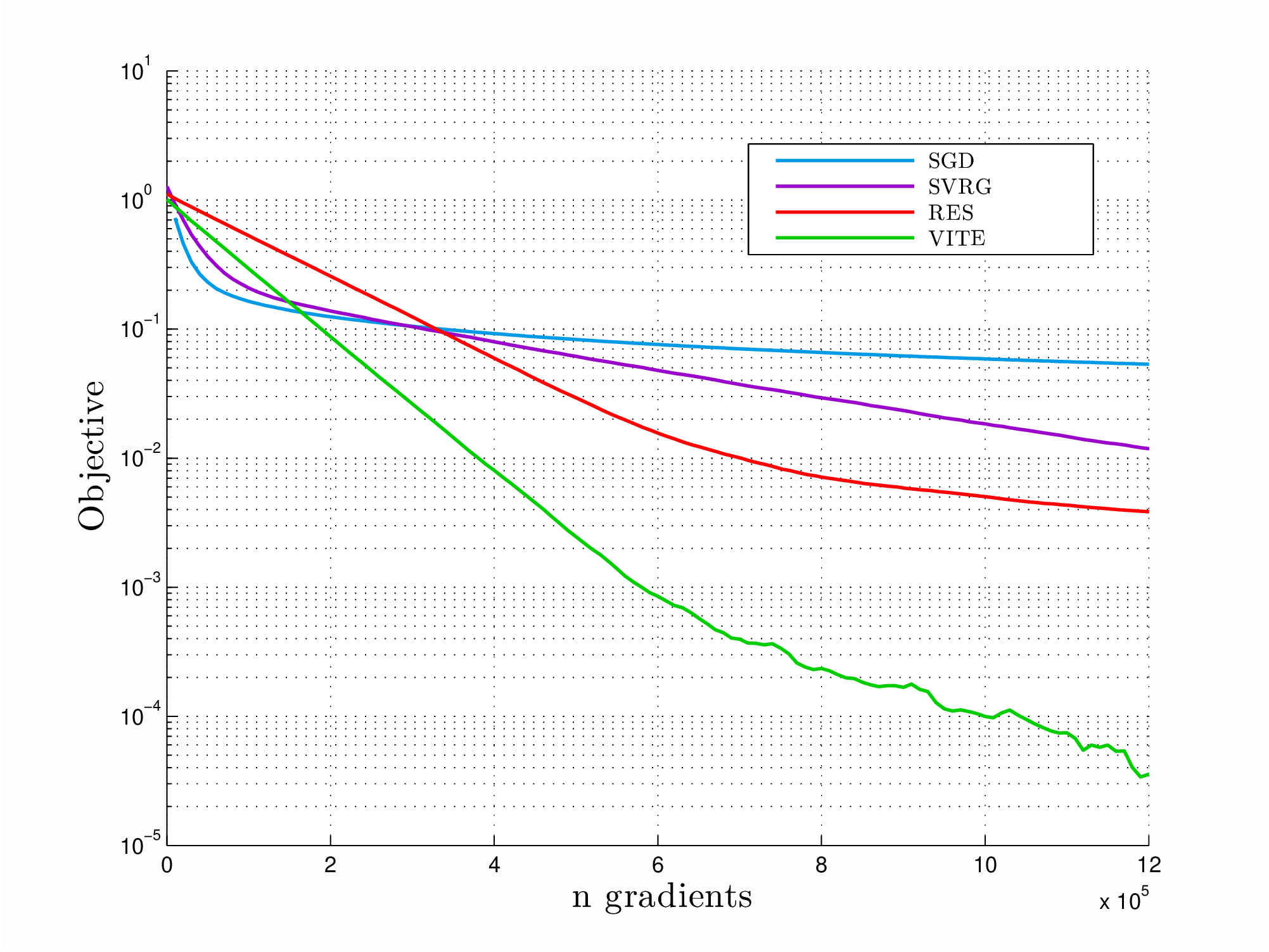} \\
            (a) {\sc Cov} &
            (b) {\sc Adult} &
            (c) {\sc Ijcnn}
            \\
	  \end{tabular}
          \caption{\footnotesize{{\it Comparison of RES and {\sc Vite} (trained with
            the best performing parameters) against SGD and SVRG. The reduction
            in variance for {\sc Vite} is faster than SGD or RES
            which typically lead to faster convergence.}}}
          \label{fig:results_sgd}
	\end{center}
	\vspace{0.7cm}
\end{figure*}

Finally, we compared {\sc Vite} against
SGD, RES and SVRG~\cite{johnson2013, konevcny2013}. A critical factor in the performance of
SGD is the selection of the step-size. 
We use the step-size given in
Eq.~\ref{eq:step_size}{\it b} and pick the parameters $T_0$ and
$\eta_0$ by performing cross-validation over $T_0 = \{1, 10, 10^2,
\dots, 10^4\}$ and $\eta_0 = \{10^{-1}, \dots, 10^{-5}\}$. Although it is a quasi-Newton method, RES also requires a decaying step-size and so the same selection process was performed. For SVRG and {\sc Vite},
we used a \emph{constant} step size chosen in the same interval as $\eta_0$.  For SVRG and {\sc Vite} we used the same size subset, $\C$ to compute $\mutil$. 
Fig.~\ref{fig:results_sgd} shows the
objective value of each method in log scale. Although RES and SVRG are superior to SGD, neither clearly outperforms the other. On the other hand, we observe that {\sc Vite} consistently
converges faster than both RES and SVRG. This demonstrates that the combination of second order information \emph{and} variance reduction is beneficial for fast convergence.

\section{Conclusion}
We have shown that stochastic variants of BFGS can be made more robust
to the effects of noisy stochastic gradients using variance reduction. We introduced {\sc Vite} and
showed that
it obtains a geometric convergence rate for smooth convex functions -- to our knowledge the first stochastic Quasi-Newton algorithm with this property. We have shown experimentally that {\sc Vite} outperforms both variance reduced SGD and
stochastic BFGS. The theoretical analysis we present is quite general and additionally only requires that the bound on the eigenvalues of the inverse Hessian matrix in \eqref{eq:bound_inv_hessian} holds. Therefore, the variance reduced framework we propose can be extended to other quasi-Newton methods, including the widely used L-BFGS and {\sc AdaGrad} \cite{duchi:2011} algorithms. 
Finally, an important open question is how to bridge the gap between the theoretical and empirical results. Specifically, whether it is possible to obtain better convergence rates for stochastic BFGS algorithms which match the improvement we have demonstrated over SVRG.

\newpage

\section{Appendix}

\subsection{Proof of Lemma 2}


\begin{align}
\E & \sqnorm{\vt} = \E \sqnorm{\nabla f_i(\w_{t}) - \nabla f_i(\wtil) + \nabla f(\wtil)} \nonumber \\
&\leq 2 \E \sqnorm{\nabla f_i(\w_{t}) - \nabla f_i(\wstar)} \nonumber \\
& + 2 \E \sqnorm{(\nabla f_i(\wtil) - \nabla f_i(\wstar)) - \nabla f(\wtil)} \nonumber \\
&= 2 \E \sqnorm{\nabla f_i(\w_{t}) - \nabla f_i(\wstar)} \nonumber \\
&+ 2 \E \sqnorm{(\nabla f_i(\wtil) - \nabla f_i(\wstar)) - (\nabla f(\wtil) - \nabla f(\wstar))} \nonumber \\
&\leq 2 \E \sqnorm{\nabla f_i(\w_{t}) - \nabla f_i(\wstar)} \nonumber \\
&+ 2 \E \sqnorm{\nabla f_i(\wtil) - \nabla f_i(\wstar)} \nonumber \\
&\leq 4L (f(\w_{t}) - f(\wstar) + f(\wtil) - f(\wstar))
\label{eq:proof_bound_vk}
\end{align}

The second inequality uses $\E \sqnorm{\xi - \E \xi} = \E \sqnorm{\xi} - \sqnorm{\E \xi} \leq \E \sqnorm{\xi}$ for any random vector $\xi$.
\\

The last inequality uses the following inequality derived from the fact that $f_i$ is a Lipschitz function:
$$\E \sqnorm{\nabla f_i(\wstar) - \nabla f_i(\w_t)} \leq 2L (f(\w_t) - f(\wstar)).$$ \qed

\subsection{Selection of the parameter $|\A|$.}
\begin{figure*}[h!]
	\begin{center}
          \begin{tabular}{@{}c@{\hspace{1mm}}c@{\hspace{1mm}}c@{}}
            \includegraphics[trim = 15 15 15 15, clip, width=0.3\linewidth]{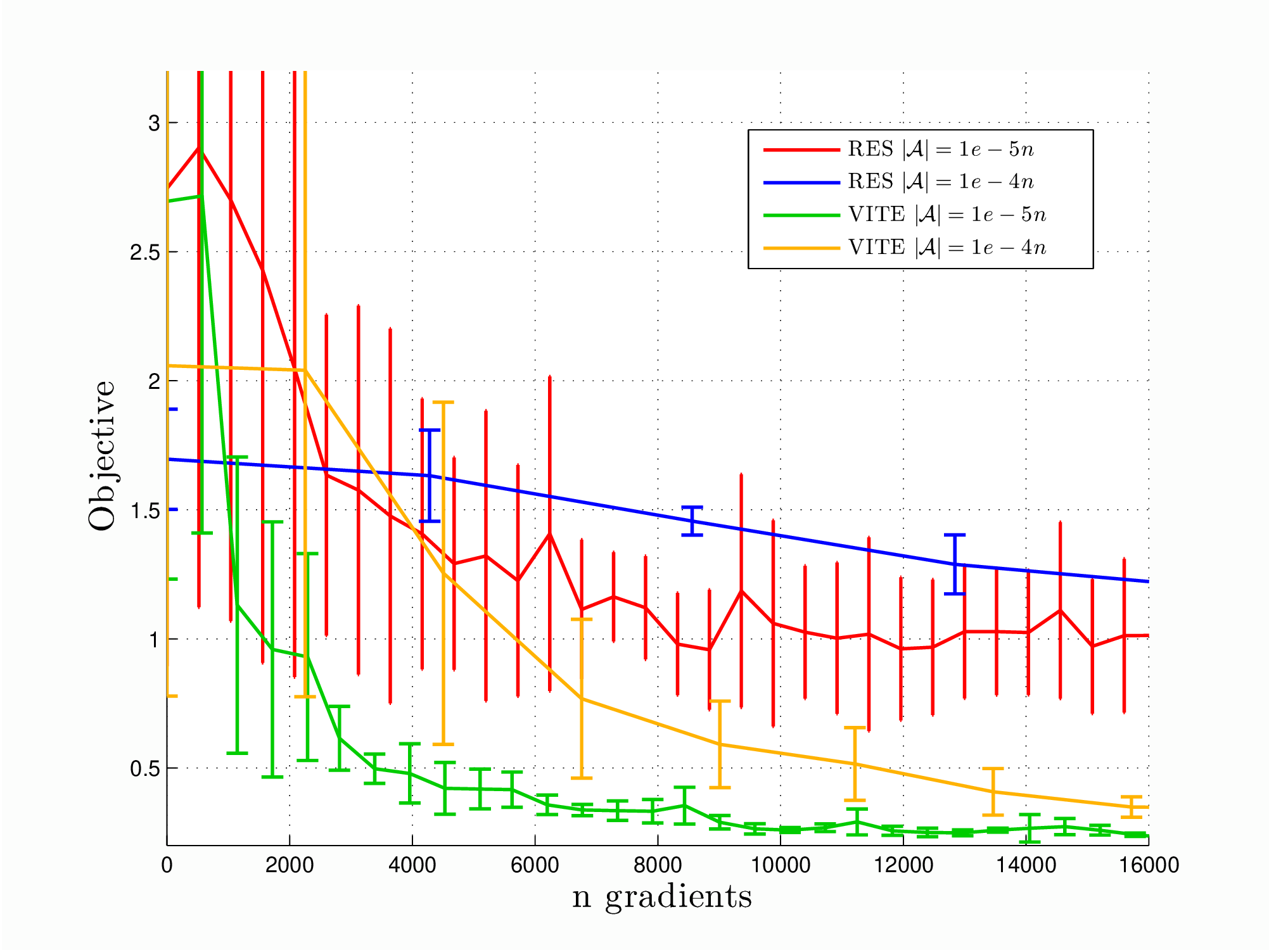} &
            \includegraphics[trim = 15 15 15 15, clip, width=0.3\linewidth]{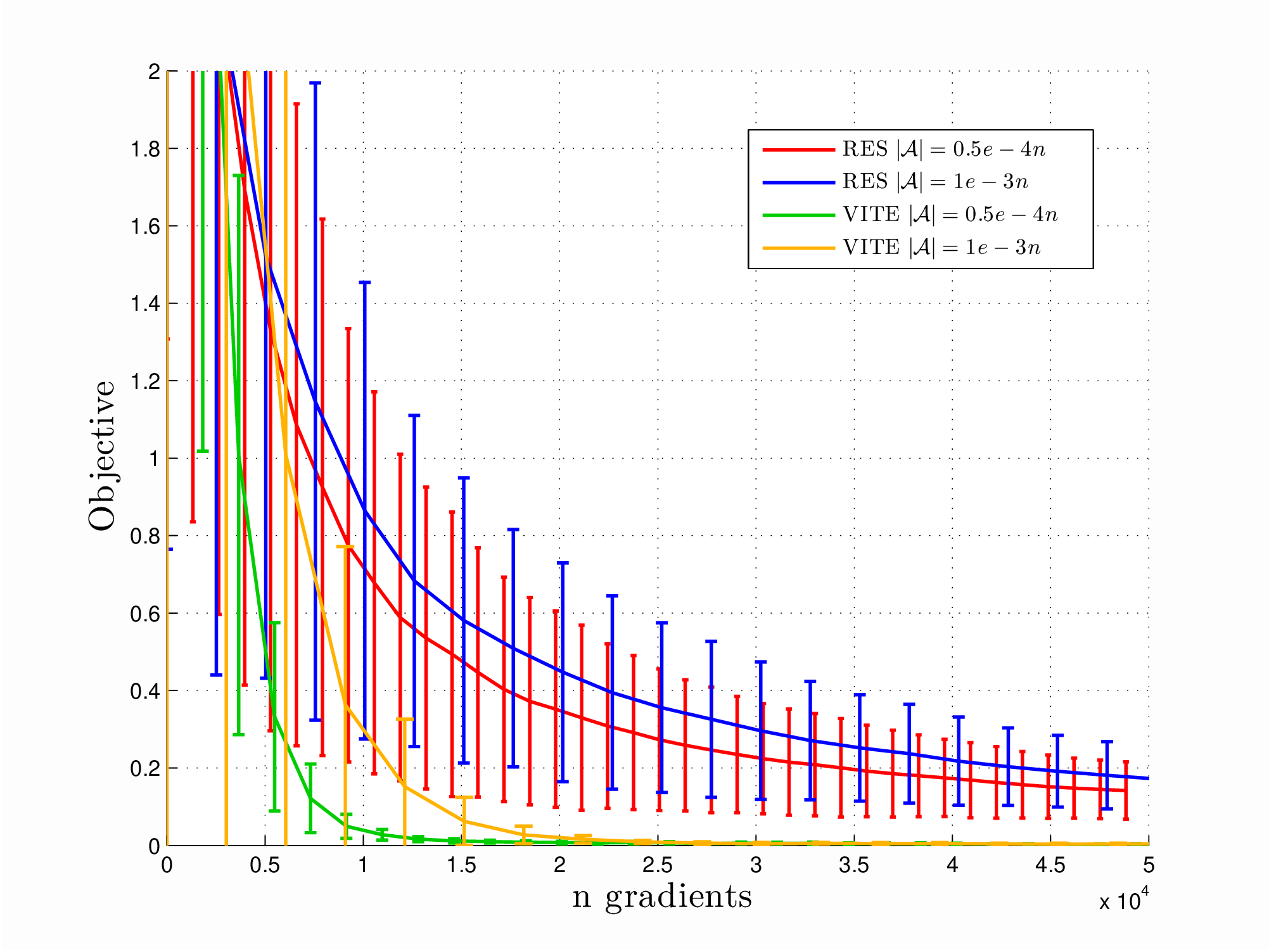} &
            \includegraphics[trim = 15 15 15 15, clip, width=0.3\linewidth]{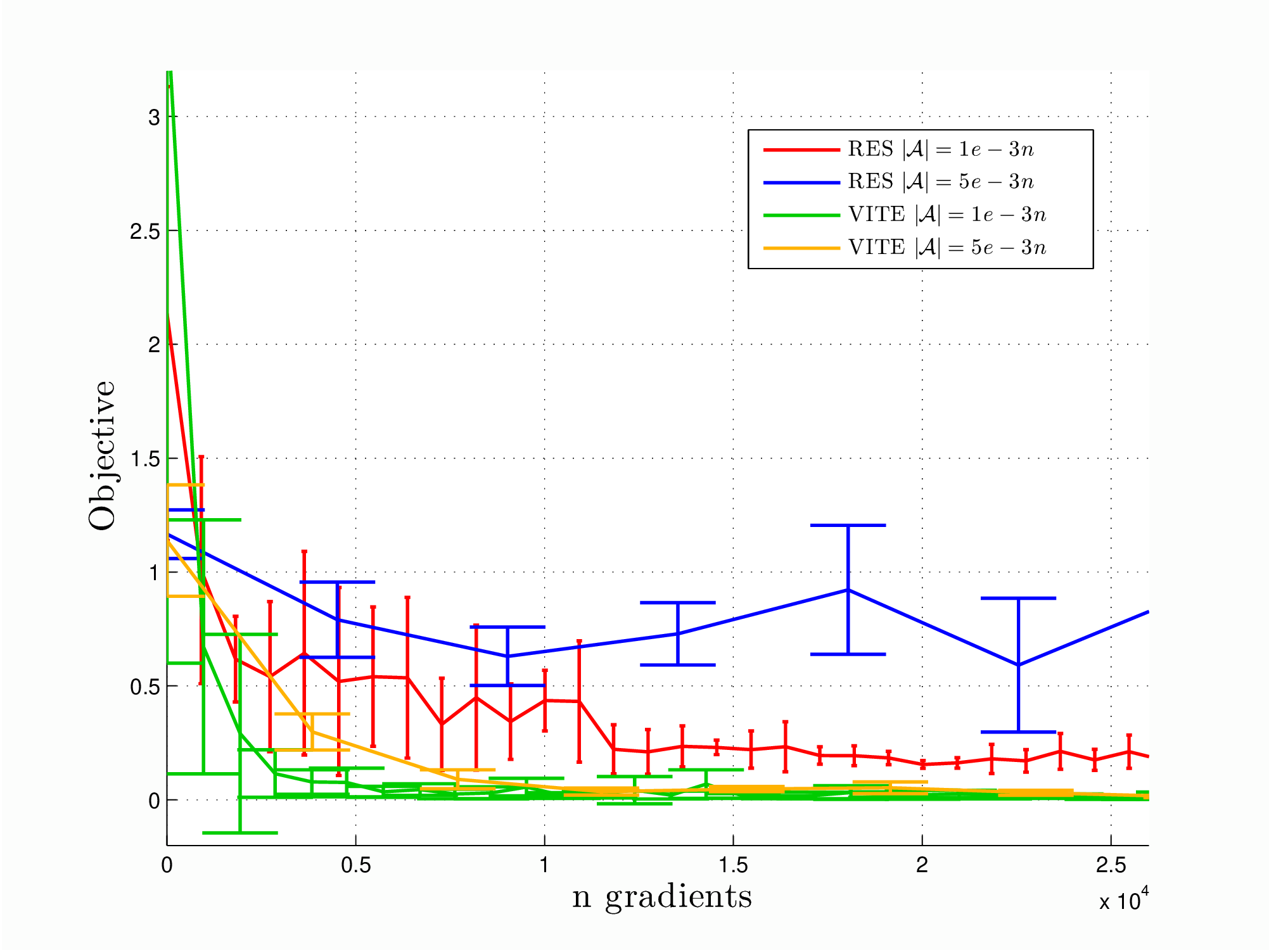} \\
            (a) {\sc Cov} &
            (b) {\sc Adult} &
            (c) {\sc Ijcnn}
            \\
	  \end{tabular}
          \vspace{2mm}
          \caption{Evolution of the objective value of RES and {\sc
              VITE} for different values of $|\A|$. We can see that
            the lowest value of $|\A|$ performs better, which
            indicates than there is no gain at increasing this value
            passed a certain cut-off value.\label{fig:results_A}}
	\end{center}
\end{figure*}

Figure~\ref{fig:results_A} shows the effect of the set $\A$, used to estimate the inverse Hessian, on the convergence of RES
and {\sc Vite}. We show results for $|\A|=\{0.00001, ~ 0.0001 \}\times n$. Firstly we see that better performance is obtained for both methods for the smaller value of $|\A|$. By increasing
$|\A|$, the penalty paid in terms of gradient evaluations outweighs
the gain in terms of better curvature estimates and so convergence is slower. A similar
observation was made in~\cite{mokhtari2014b}. However, we also observe that {\sc Vite} always outperforms RES for all combinations of $|\A|$.

\bibliographystyle{abbrv}
\bibliography{online_newton_full}

\end{document}